\documentclass[]{l4dc2027}

\newcommand{\R}{\mathbb{R}}
\newcommand{\N}{\mathbb{N}}
\newcommand{\E}{\mathbb{E}}

\DeclareMathOperator*{\argmin}{arg\,min}

\newcommand{\defe}{\mathrel{\mathop:}=}

\newcommand{\fL}{f^{L}}                     
\newcommand{\Proj}{\Pi_{[0,1]^n}}           

\newcommand{\Regret}{\mathrm{Regret}}
\newcommand{\DRegret}{\text{D-}\mathrm{Regret}}
\newcommand{\PN}{P^{\ast}_{N}}                 

\newcommand{\norm}[1]{\left\lVert #1 \right\rVert}

\newcommand{\inner}[2]{\left\langle #1, #2 \right\rangle}

\newenvironment{proofof}{\noindent {\textbf{Proof of~}}}{\hfill \hspace*{1pt} \hfill $\square$}

\usepackage{times}
\usepackage{booktabs}

\title[Bandit Online Submodular Minimisation]{%
Bandit Online Submodular Minimisation Using Zeroth-Order Oracles}
\usepackage{times}

\author{%
 \Name{Amir Ali Farzin} \Email{amirali.farzin@anu.edu.au}\\
 \addr School of Engineering, Australian National University
 \AND
 \Name{Philipp Braun} \Email{philipp.braun@anu.edu.au}\\
 \addr School of Engineering, Australian National University
 \AND
 \Name{Iman Shames} \Email{iman.shames@unimelb.edu.au}\\
 \addr Department of Electrical and Electronic Engineering,
 University of Melbourne
}

\begin{document}

\maketitle

\begin{abstract}%
We consider online submodular minimisation under bandit feedback, where at
each round a decision maker plays a subset and observes only its cost,
while the cost functions, and hence their minimiser, change over time.
Existing bandit methods only guarantee static regret, which compares
against a single fixed subset and is a weak benchmark when the minimiser
moves. This paper provides, to the best of our knowledge, the first method
with guarantees for tracking a moving minimiser in this setting. We
propose a zeroth-order method built on a two-point oracle for the
Lov\'asz extension whose two evaluations are both subsets that can be
played. The perturbed and base points are rounded by a common random
threshold, and drawing the smoothing direction from the unit sphere makes
the base evaluation a zero-mean control variate, so the oracle is an
unbiased estimator of the smoothed gradient. We establish dynamic regret
bounds of order $O\big((1+\sqrt{P^\ast_N})N^{3/4}\big)$ in general and
$O\big((1+\sqrt{P^\ast_N})N^{2/3}\big)$ under a sufficiently small drift,
where $P^\ast_N$ is the path length of the minimiser sequence, together with
the corresponding $O(N^{3/4})$ and $O(N^{2/3})$ static regret bounds.
Numerically, the method attains lower regret than the algorithm of
\citet{hazan2012online}, with the advantage growing with the path length.
\end{abstract}


\section{Introduction}
\label{sec:intro}


Submodular set functions formalise diminishing marginal returns and play a central role in combinatorial optimisation and machine learning \citep{edmonds2003submodular, fujishige2005submodular, krause2010submodular, jegelka2013reflection,farzin2026solving}. Their minimisation is solvable in polynomial time \citep{grotschel1981ellipsoid}. We refer to \citet{iwata2008submodular} and \citet{mccormick2005submodular} for surveys. The Lovász extension \citep{lovasz1983submodular} converts this discrete problem into convex optimisation over the unit hypercube. Although a subgradient can be computed via a greedy chain using \(n\) value-oracle queries \citep{hazan2012online}, this approach is not directly available under bandit feedback.

We consider the following online  {submodular function
 minimisation (SFM)} problem. At each \emph{round} $k=0,1,\dots,N$, a decision maker selects
(\emph{plays}) a subset
$S_k\subset {\{1,\dots,n\}}$ and incurs the cost $f_k(S_k)$, where
$f_0,\dots,f_N$ are submodular functions  {with values in $[-M,M]$, for a
known constant $M>0$,} that are fixed in advance but unknown to the decision
maker. Under \emph{bandit feedback}, the scalar $f_k(S_k)$ is the only
information revealed.  {This arises whenever a cost can only be observed
by deploying the subset, as in online sensor or actuator selection.} This
setting differs from \emph{offline} SFM, in which a single function is
minimised using arbitrarily many value queries, and from online SFM under
\emph{full information}, in which $f_k$ may be queried at any subset once
$S_k$ has been played. 
Performance is measured by the \emph{regret},  {i.e., the total cost
incurred minus that of a benchmark, which is either the best single subset
(\emph{static} regret) or the minimiser of each round (\emph{dynamic}
regret).}



Under full information, \citet{hazan2012online} and \citet{jegelka2011online}
establish an $O(\sqrt N)$ static regret bound. Under bandit feedback,
\citet[Thm.~16]{hazan2012online} give an algorithm attaining an
$O(MnN^{2/3})$ static regret bound, and \citet{ito22a}
obtains gap-dependent bounds.  
To the best of our knowledge, no dynamic
regret guarantee is available for online SFM under bandit feedback.
Under bandit feedback the greedy subgradient scheme is unavailable, since it
requires $n$ evaluations of the same cost function, of which only one can
be played. Zeroth-order (ZO) methods, which rely only on function
evaluations \citep{audet2017introduction}, are therefore natural candidates
 {to solve the problem.}  { Gaussian and unit ball smoothing schemes have
been studied for a range of settings \citep{nesterov_random_2017,
flaxman2004online, shames2019online, farzin2024minimisation,
farzin2025minimisation, farzin2025minmax}. A direct application fails,
however. Two-point ZO oracles need the value of the Lov\'asz extension at
interior points of the hypercube, and such a value is a convex combination
of the costs of several subsets, which is not observable. Instead, we use
its probabilistic representation. Rounding a point by a uniformly random
threshold yields a subset whose expected cost equals the value of the
extension at that point. Our estimator rounds the perturbed and the base
point with a \emph{common} threshold.} Since this requires a threshold range
fixed in advance, and hence a bounded perturbation, we use unit ball smoothing
\citep{flaxman2004online} rather than Gaussian smoothing
\citep{nesterov_random_2017}.

The bandit algorithm of \citet{hazan2012online} writes the current iterate
as a convex combination of the characteristic vectors of the sets in an
associated maximal chain, and plays one of these sets, drawn with a
probability that depends on its coefficient.  {The resulting
importance-weighted estimate of the greedy subgradient is sparse, and each
iteration modifies at most one coordinate of the iterate
\citep[Alg.~3]{hazan2012online}. Our estimator is aligned with a direction
drawn from the unit sphere and therefore updates every coordinate at every
iteration.} When the minimiser moves, a method that alters a single
coordinate per  {update}
requires at least as many
 {updates} as there are coordinates in which the iterate and the
minimiser differ.  {Consistently, in Section~\ref{sec:numerics} the
dynamic regret of our method is close to that of \citet{hazan2012online}
when the minimiser is fixed and increasingly lower as it moves more often.
}

\paragraph{Contributions.}
For the bandit problem above, the contributions of this paper are as
follows.
\begin{itemize}
  \item We propose a ZO oracle for the Lov\'asz extension in which
  the perturbed and the base point are rounded by a common random
  threshold, so that both evaluations can be played. The base evaluation
  acts as a zero-mean control variate, and the oracle is an unbiased
  estimator of the smoothed gradient with no assumption relating the two
  observed costs (Lemma~\ref{lem:unbiased}). We show that drawing the smoothing direction from the unit sphere
  is what permits a fixed threshold range to work.
  \item We derive, to the best of our knowledge, the first dynamic regret
  bounds for online SFM under bandit feedback, expressed through the path
  length $\PN$ of the minimiser sequence. They are of order
  $O\big((1+\sqrt{\PN})N^{3/4}\big)$ in general and
  $O\big((1+\sqrt{\PN})N^{2/3}\big)$ under the same drift condition
  (Theorem~\ref{thm:dynamic}). Also, we establish an $O(N^{3/4})$ static regret bound in general and
  $O(N^{2/3})$ when consecutive cost functions differ by a sufficiently
  small amount. 
  \item Numerically, the method attains lower regret than the algorithm of
  \citet{hazan2012online}, with the advantage growing with the path length,
  and an ablation isolates the role of the base evaluation
  (Section~\ref{sec:numerics}).
\end{itemize}


\paragraph{Notation.}
 The Euclidean norm is denoted by $\norm{\cdot}$ and $\inner{x}{y}=x^\top y$  {denotes the inner product}. We write
$[n]=\{1,\dots,n\}$, $2^{[n]}$ for the power set of $[n]$, and $x(i)$ for the
$i$th component of $x$. For $S\subset[n]$, $\chi_S\in\{0,1\}^n$ is its
characteristic vector  {($\chi_S(i)=1$ if $i\in S$ and $\chi_S(i)=0$ otherwise)}, 
and we use $f(S)$ and $f(\chi_S)$ interchangeably.
 {The function} $\Proj {(\cdot)}$ denotes the projection onto $[0,1]^n$, acting componentwise as
$\Proj(x)(i)=\max\{0,\min\{1,x(i)\}\}$. Finally $\mathcal{N}(0,I)$ is the
standard Gaussian on $\R^n$ and $U[a,b]$ the uniform distribution on $[a,b]$.  {For $w\in\R^n$ and $\tau\in\R$, $S_\tau(w)\defe\{i\in[n]:w(i)>\tau\}$
denotes the level set of $w$ at threshold $\tau$.}

\section{Preliminaries}\label{sec:prelim}
In this section, we present the necessary background material of online optimisation problems, submodular functions, the Lov\'asz extension and unit ball smoothing.

\subsection{Problem Formulation}\label{subsec:problem}

We study the online minimisation of a sequence of submodular set functions,
\begin{equation}\label{eq:mainp}
  \min_{S_k\subset[n]} f_k(S_k), \qquad k=0,1,\dots,N,
\end{equation}
 {where each $f_k:2^{[n]}\to[-M,M]$ is submodular, for a constant $M>0$ independent of $N.$}
That is, each $f_k$ satisfies     $f_k(S \cup \{i\}) -  {f}_k(S) \geq  f_k(T \cup \{i\}) -  f_k(T)$
for all $S \subset T \subset [n]$, $i \notin T$, as made precise in
Definition~\ref{def:sub} in the following. Moreover, we assume that $f_k$  
satisfies $f_k(\emptyset)=0$ 
without loss of generality, since subtracting the constant $f_k(\emptyset)$
alters neither the minimisers nor submodularity. Problem~\eqref{eq:mainp} is
an online optimisation problem in the following sense. In 
round $k$ the decision maker must commit to a set $S_k$ and the 
functions
$f_k$ are fixed and unknown to the decision maker in advance. 
The sequence $\{f_k\}$ does not depend on the
decisions of the decision maker.

We distinguish two feedback models.
Under \emph{full information}, after playing $S_k$ the decision maker may query $f_k$ at arbitrarily many sets \citep{jegelka2011online,cardoso2019differentially}.
Under \emph{bandit feedback}, considered here, only the scalar
$f_k(S_k)$ is observed and $f_k$ is not accessible.   {We refer to any subset of $[n]$ as an \emph{admissible action}.}
Performance is measured by static and dynamic regrets, as defined below. 

\begin{definition}[Static and dynamic regret]\label{def:regret}
The \emph{static} and \emph{dynamic} regret of the decision maker over $N+1$
rounds are, respectively,
\begin{equation}\label{eq:regrets}
  \Regret_N \defe \sum_{k=0}^{N} f_k(S_k)
    - \min_{S\subset[n]}\sum_{k=0}^{N} f_k(S),
  \quad
  \DRegret_N \defe \sum_{k=0}^{N} f_k(S_k)
    - \sum_{k=0}^{N}\min_{S\subset[n]} f_k(S).
\end{equation}

\end{definition}
The comparator in $\Regret_N$ is a single set held fixed over the horizon,
whereas $\DRegret_N$ compares against the per-round minimiser. Consequently  {the inequality}
$\DRegret_N\ge\Regret_N$  {holds.} 
Dynamic regret depends on how fast the comparator moves. The standard 
quantity  {in this context} is
the path length \citep{zinkevich2003online, hazan2016introduction}.

\begin{definition}[Path length]\label{def:pathlength}
Let $S^\ast_k\in\argmin_{S\subset[n]}f_k(S)$  {be any selection of minimisers}. 
The path
length of the minimiser sequence is
$\PN\defe\sum_{k=0}^{N-1}\|\chi_{S^\ast_{k+1}}-\chi_{S^\ast_k}\|$.
\end{definition}


Finally, following \citet{shames2019online}, we allow the cost function to
change between 
two evaluations required by a two-point oracle, i.e., an oracle returning an estimate of the gradient of a smoothed function of $f_k$ based on two function evaluations.
The specific two-point oracle used in this work is introduced in Section
\ref{subsec:oracle}.
We therefore
work with a family $\{f_k\}$ indexed by
$k\in\N\cup\{j+\tfrac12: j\in\N\}=\{0,\tfrac12,1,\tfrac32,\dots\}$ and
abbreviate $k+\tfrac12$ by $k^{+}$. 
Submodularity, $f_{k^+}(\emptyset)=0$ and $f_{k^+}:2^{[n]}\to[-M,M]$ hold
for every index in this family.  {Each round $k$ thus consists of two plays, against $f_k$ and $f_{k^+}$,
followed by one update of the iterate.}

\subsection{Submodular Functions and the Lov\'asz Extension}\label{subsec:lovasz}

\begin{definition}\label{def:sub}
    A function $ {f} : 2^{[n]} \to \mathbb{R}$ is submodular if, for all $S \subset T \subset [n]$ and any element $i \notin T$, 
\(
     {f}(S \cup \{i\}) -  {f}(S) \;\;\ge\;\;  {f}(T \cup \{i\}) -  {f}(T).
\)  
\end{definition}

The \emph{Lov\'asz extension} provides a continuous, convex extension of a
submodular function $f:2^{[n]}\to\R$  {from $2^{[n]}$} to 
$\R^n$. Since the analysis
of Section~\ref{sec:main} evaluates  {the Lov\'asz extension} 
at points outside the unit hypercube, we
state it in this generality and then record the two specialisations. 

\begin{definition}[Lov\'asz extension, \citealp{bach2013learning}, Def.~3.1]
\label{def:lovae}
Let $f:2^{[n]}\to\R$ satisfy $f(\emptyset)=0$. For $w\in\R^n$, order the
components in decreasing order $w(j_1)\ge\cdots\ge w(j_n)$, where
$(j_1,\dots,j_n)$ is a permutation of $[n]$. The Lov\'asz extension
$f^L:\R^n\to\R$ of $f$ is
\begin{equation}\label{eq:lovasz_sort}
  f^L(w)=\sum_{\ell=1}^{n} w(j_\ell)\big[f(\{j_1,\dots,j_\ell\})
  -f(\{j_1,\dots,j_{\ell-1}\})\big],
\end{equation}
or equivalently, in integral form \citep[eq.~(3.4)]{bach2013learning},
\begin{equation}\label{eq:lovasz_choquet}
  f^L(w)=\int_{0}^{+\infty} f(\{i:w(i)\ge z\})\,\mathrm{d}z
  +\int_{-\infty}^{0}\big[f(\{i:w(i)\ge z\})-f([n])\big]\,\mathrm{d}z .
\end{equation}
\end{definition}

On the unit hypercube it
reduces to the familiar combinatorial and probabilistic forms.

\begin{remark}[Restriction to {$[0,1]^n$}, \citealp{hazan2012online}, Def.~5]
\label{rem:lovae}
Let $x\in[0,1]^n$ and let $A_0\subsetneq A_1\subsetneq\cdots\subsetneq A_p\subset[n]$,
$p\le n$, be the chain of subsets for which $x$ can be written as the convex
combination $x=\sum_{i=0}^{p}\lambda_i\chi_{A_i}$ with $\lambda_i>0$ and
$\sum_{i=0}^{p}\lambda_i=1$. Then \eqref{eq:lovasz_sort} coincides with
\(
  f^L(x)=\sum_{i=0}^{p}\lambda_i f(A_i).
\)
Moreover, sampling a threshold $\tau\in[0,1]$ uniformly at random and considering
the level set $S_\tau(x)$,  {it holds that}
\begin{align}\label{eq:eqle2}
  f^L(x)=\E_\tau\big[f(S_\tau(x))\big] \quad \text{ and } \quad f^L(\chi_S)=f(S) \ \  \forall \ S\subset[n].   
\end{align}
\end{remark}

Representation \eqref{eq:eqle2} is the mechanism on which 
 the algorithm {in} 
Section~\ref{sec:main} rests. The level set $S_\tau(x)$ is an element of
$2^{[n]}$, hence an admissible action, and its expected cost equals the value
of the continuous surrogate at the point being rounded. The next remark extends
this to arbitrary points of $\R^n$, at the price of a threshold drawn from a
wider range.

\begin{remark}[Restriction to {$[a,b]^n$}]
\label{rem:extended}
Let $w\in\R^n$ and let $[a,b]$ satisfy $a\le\min_i w(i)$, $\max_i w(i)\le b$
and $a\le 0\le b$. Then the integrands in \eqref{eq:lovasz_choquet} vanish
outside $[a,b]$, since $\{i:w(i)\ge z\}=[n]$ for $z<a$ and
$\{i:w(i)\ge z\}=\emptyset$ for $z>b$, and
\begin{equation}\label{eq:lovasz_boxed}
  f^L(w)=\int_{a}^{b} f(S_z(w))\,\mathrm{d}z + a\,f([n])
  =(b-a)\,\E_{\tau\sim U[a,b]}\!\big[f(S_\tau(w))\big] + a\,f([n]).
\end{equation}
Taking $[a,b]=[0,1]$ for $w\in[0,1]^n$ recovers \eqref{eq:eqle2}.
\end{remark}
\begin{lemma}[\citealp{lovasz1983submodular,fujishige2005submodular,bach2013learning,ando2002k}]
\label{lem:lovasz_props}
Let $f:2^{[n]}\to\R$ be submodular with Lov\'asz extension $\fL$. Then
\emph{(i)} $\fL$ is convex on $\R^n$; \emph{(ii)} $\fL$ is piecewise linear;
\emph{(iii)} $\min_{S\subset[n]}f(S)=\min_{x\in\{0,1\}^n}\fL(x)
=\min_{x\in[0,1]^n}\fL(x)$.
\end{lemma}
By Lemma~\ref{lem:lovasz_props}(iii), each round of \eqref{eq:mainp} is
equivalent to the continuous convex problem
\begin{equation}\label{eq:mainp_cont}
  \min_{x\in[0,1]^n}\fL_k(x).
\end{equation}
 {The function  $\fL$ is non-smooth and its subgradient requires $n$ function queries
\citep[Prop.~7]{hazan2012online} and is therefore unavailable under bandit
feedback.}
\citet[Sec.~4]{hazan2012online} proposed an adaptation of how to use the subgradient formulation for the bandit setting.

\subsection{Unit Ball Smoothing and Zeroth-Order Random Oracles}\label{subsec:sphere}

For a continuous $F:\R^n\to\R$ and smoothing parameter $\mu>0$, the
\emph{unit-ball-smoothed} version of $F$ is
$
  F_\mu(x)\defe\E_{v\sim\mathbb{B}}\big[F(x+\mu v)\big],
$
where $v$ is drawn uniformly from the unit ball
$\mathbb{B}=\{v\in\R^n:\norm{v}\le 1\}$. It is a classical fact
\citep{flaxman2004online} that $F_\mu$ is differentiable whether or not $F$ is and $F_\mu$ is convex whenever $F$ is. Its gradient admits the
representation
\begin{equation}\label{eq:sphere_grad}
  \nabla F_\mu(x)=\frac{n}{\mu}\,
  \E_{u\sim\mathbb{S}}\big[F(x+\mu u)\,u\big],
\end{equation}
where $u$ is drawn uniformly from the unit sphere
$\mathbb{S}=\{u\in\R^n:\norm{u}=1\}$. Since $\E_u[u]=0$, subtracting from
$F(x+\mu u)$ any quantity independent of $u$ leaves \eqref{eq:sphere_grad}
unchanged. We define the \emph{two-point} random oracle
\begin{equation}\label{eq:sphere_twopoint}
  g_\mu(x)\defe\frac{n}{\mu}\big(F(x+\mu u)-F(x)\big)u.
\end{equation}
The subtracted term is thus a zero-mean \emph{control variate}.
Every coordinate of
the perturbed point satisfies $(x+\mu u)(i)\in[-\mu,\,1+\mu]$ whenever
$x\in[0,1]^n$. Consequently
representation~\eqref{eq:lovasz_boxed} applies with a range that does not depend
on $u$. This is precisely the property required for the base evaluation in
\eqref{eq:sphere_twopoint} to act as a zero-mean control variate under a
$u$-independent threshold. The corresponding difficulty for Gaussian smoothing,
whose perturbations are unbounded, is discussed in
Remark~\ref{rem:gaussian_fail}.
Applying \eqref{eq:sphere_twopoint} with $F=  f^L_k$ still requires a value
of $  f^L_k$ at two points, which by \eqref{eq:lovasz_sort} costs $2n$
value-oracle calls per direction. In
the offline, full-information setting this renders the ZO approach
uncompetitive \citep{farzin2025sfmzo}. Under bandit feedback, $  f^L_k(x)$ at an arbitrary $x$ is not the cost of any admissible
action, so a direct evaluation is unavailable. 
Section~\ref{sec:main} shows
that combining \eqref{eq:sphere_twopoint} with the threshold
representation~\eqref{eq:lovasz_boxed} removes this obstruction.

\section{Main Results}
Here, we present the main results of the paper. All 
proofs can be found in Appendix~\ref{app:proof}.
\label{sec:main}
\subsection{ZO Oracle and Algorithm}
\label{subsec:oracle}

 {In this section we} 
specialise the unit ball smoothing oracle of Section~\ref{subsec:sphere} to
the Lov\'asz extension under bandit feedback. Fix  {a smoothing parameter $\mu>0$ and} an iterate $x\in[0,1]^n$,
draw $u\sim\mathbb{S}$ and $\tau\sim U[a,b]$ with the fixed range
\(
  [a,b]\defe[-\mu,\,1+\mu].
\)
The two-point threshold oracle is
\begin{equation}\label{eq:oracle}
  \hat g_\mu(x)\defe\frac{n(b-a)}{\mu}
  \big(  f_k(S_\tau(x+\mu u)) -   f_{k^+}(S_\tau(x))\big)\,u,
  \qquad b-a=1+2\mu.
\end{equation}
 {The two arguments $S_\tau(x+\mu u)$ and $S_\tau(x)$ are admissible
actions, which we call the \emph{exploration play} and the \emph{base
play}.}
In
the online setting the two evaluations occur at consecutive plays and may
therefore involve distinct cost functions $  f_k$ and $  f_{k^+}$, in the
sense of Section~\ref{subsec:problem}.
Algorithm~\ref{alg:main} performs projected descent along $\hat{g}_k$.
We 
show that $\hat g_\mu$ is an unbiased
estimator of the gradient of the unit-ball-smoothed function
and bound its second moment.
\begin{algorithm2e}[t]
\caption{Bandit Two-Point Threshold Zeroth-Order Descent}
\label{alg:main}
\KwIn{$x_0 \in [0,1]^n$, step size $h>0$, smoothing parameter $\mu>0$,
      horizon $N,$ $b-a=1+2\mu$}
\For{$k = 0,\dots,N$}{
  sample $u_k \sim \mathbb{S}$ and $\tau_k \sim U[-\mu,\,1+\mu]$\;
  play $S_{\tau_k}(x_k + \mu u_k)$ and observe $c_k =   f_k(S_{\tau_k}(x_k+\mu u_k))$ \tcp*{exploration play}
  play $S_{\tau_k}(x_k)$ and observe $c_{k^+} =   f_{k^+}(S_{\tau_k}(x_k))$ \tcp*{base play}
  $\hat g_k = \dfrac{n(b-a)}{\mu}\,(c_k - c_{k^+})\,u_k$\;
  $x_{k+1} = \Proj\!\big( x_k - h\,\hat g_k \big)$\;
}
\Return $x_k$ for $k = 0,\dots,N$
\end{algorithm2e}

\begin{lemma}[Unbiasedness and second-moment bounds]\label{lem:unbiased}
Let $f_k$ be submodular with Lov\'asz extension
$f^L_k$, and let $f^L_{\mu,k}$ denote its unit ball smoothing  {with parameter $\mu>0$}. Fix $x\in[0,1]^n$, let $u\sim\mathbb{S}$ and
$\tau\sim U[-\mu,1+\mu]$ be independent, and let $\hat g_\mu(x)$ be the oracle
\eqref{eq:oracle}. 
Then
\begin{align}
&  \E_{u,\tau}\big[\hat g_\mu(x)\big] = \nabla f^L_{\mu,k}(x) \qquad\text{and} \qquad
    \E_{u,\tau}\big[\norm{\hat g_\mu(x)}^2\big]
    \;\le\; \frac{4 M^2 n^2 (1+2\mu)^2}{\mu^2}.  \label{eq:unbiased}
\end{align}

Moreover, if  $ |f_k(S)-f_{k^+}(S)|\le V \ \  \forall \  S\subset[n]$,  then
\begin{align}
 \E_{u,\tau}\big[\norm{\hat g_\mu(x)}^2\big]
     \;\le\; \frac{n^2(1+2\mu)^2}{\mu^2}
     \Big( \frac{8M^2\mu\sqrt n}{1+2\mu} + 2V^2 \Big). 
    \label{eq:secondmoment_V}
  \end{align}

\end{lemma}

 {Lemma~\ref{lem:unbiased} relies on $\tau$ being independent of $u$ and Remark~\ref{rem:gaussian_fail} in Appendix~\ref{app:rem} explains why other alternatives such as $u$-dependent threshold range or for Gaussian smoothing may fail.}

\subsection{Static Regret}
\label{subsec:static}

We now bound the static regret \eqref{eq:regrets} of Algorithm~\ref{alg:main}.
Throughout, $\bar L_0=\max_k L_{0,k}$ denotes a uniform Lipschitz constant of
the extensions $f^L_k$, which exists  {since, by Lemma~\ref{lem:lovasz_props} and \eqref{eq:lovasz_sort}, $f^L_k$ is piecewise
linear with coefficients bounded by $2M$ in absolute value.}
Each
iteration consists of an exploration play $S_{\tau_k}(x_k+\mu u_k)$, which is  {counted in the regret}, and a base play
$S_{\tau_k}(x_k)$, which serves as a variance-reducing probe. The index $N$ counts iterations.

\begin{theorem}[Static regret]\label{thm:static}
 {For $N\geq 0$, let}  
$x^\ast\in\argmin_{x\in[0,1]^n}\sum_{k=0}^{N} f^L_k(x)$, $\{x_k\}_{k=0}^N$ be generated
by Algorithm~\ref{alg:main} with constant step size $h_k=h>0$ and  $\mu>0,$ $S_k=S_{\tau_k}(x_k+\mu u_k)$  {and let $\bar{L}_0$ be a uniform Lipschitz constant of $f^L_k$, $k\in \{0,\ldots,N\}$. Then the following bounds are satisfied.}


\smallskip

\noindent (i) For any $N\ge 0$,
\begin{equation}\label{eq:static_general}
  \frac{1}{N+1}\sum_{k=0}^{N}
  \Big(\E\big[f^L_k(x_k)\big]-f^L_k(x^\ast)\Big)
  \;\le\;
  \frac{n}{2h(N+1)}
  \;+\; \mu \bar L_0
  \;+\; \frac{2h M^2 n^2(1+2\mu)^2}{\mu^2}.
\end{equation}
Choosing  {$\mu$, $h$ and $N$ such that}
\begin{align}
  &\mu=\frac{\sqrt{n}}{(N+1)^{1/4}} \leq 1,
  \qquad
  h=\frac{1}{6M(N+1)^{3/4}}  \qquad N+1\ge n^{2}, \label{eq:static_tuning} \\
 &\text{yields} \qquad  \E\big[\Regret_N\big]
  \;\le\; \big(2\sqrt{n}\bar L_0+ 3M\sqrt{n} + 6Mn\big)(N+1)^{3/4}. \label{eq:static_rate}
\end{align}

\noindent (ii) If $|f_k(S)-f_{k^+}(S)|\le V$ for all $S\subset[n]$
and all $k\in \{0,\ldots,N\}$, then for any $N\ge0$,
\begin{equation}\label{eq:static_general_V}
  \frac{1}{N+1}\!\sum_{k=0}^{N}
  \Big(\E\big[f^L_k(x_k)\big]\!-\!f^L_k(x^\ast)\Big)
  \!\le\!
  \frac{n}{2h(N+1)}
  \!+\! \mu \bar L_0
  \!+\! \frac{h n^2(1+2\mu)^2}{\mu^2}\big(V^2+\frac{4M^2\mu\sqrt n}{1+2\mu}\big).
\end{equation}
Choosing  {$\mu$, $h$ and $N$ such that}
\begin{align}
  &\mu=\frac{n}{(N+1)^{1/3}}\leq 1,
  \ \
  h=\frac{1}{4\sqrt{3}Mn^{1/4}(N+1)^{2/3}},  \ \ N+1\ge n^{3}, \quad V\le \sqrt{\tfrac{4}{3}} Mn^{1/4}\sqrt{\mu}  \label{eq:static_tuning_V} \\
& \text{yields} \qquad  \E\big[\Regret_N\big]
  \;\le\; \big(2n\bar L_0+3Mn + 4\sqrt{3}Mn^{5/4}\big)(N+1)^{2/3} . \label{eq:static_rate_V}
\end{align}
\end{theorem}

\subsection{Dynamic Regret}
\label{subsec:dynamic}

We now turn to the dynamic regret \eqref{eq:regrets}, in which the comparator
tracks the per-round minimiser. For each $k$ let
$S^\ast_k\in\argmin_{S\subset[n]}f_k(S)$ and $x^\ast_k=\chi_{S^\ast_k}$, and recall the
path length $\PN$ of
Definition~\ref{def:pathlength}.

\begin{theorem}[Dynamic regret]\label{thm:dynamic}
 {For $N\geq 0$, let} 
$\{x_k\}_{k=0}^N$ be generated by
Algorithm~\ref{alg:main} with constant step size $h_k=h>0$ and $\mu>0,$ $S_k=S_{\tau_k}(x_k+\mu u_k)$, let $D_N \defe n+3\sqrt{n}\,\PN$, 
 {and let $\bar{L}_0$ be a uniform Lipschitz constant of $f^L_k$, $k\in \{0,\ldots,N\}$. Then the following bounds are satisfied.}

\noindent (i) For any $N\ge0$,
\begin{equation}\label{eq:dyn_general}
  \frac{1}{N+1}\sum_{k=0}^{N}
  \Big(\E\big[f^L_k(x_k)\big]-f^L_k(x^\ast_k)\Big)
  \;\le\;
  \frac{D_N}{2h(N+1)}
  \;+\; \mu\bar L_0
  \;+\; \frac{2hM^2n^2(1+2\mu)^2}{\mu^2}.
\end{equation}
Choosing  {$\mu$, $h$ and $N$ such that}
\begin{align}
  \mu=\frac{\sqrt n}{(N+1)^{1/4}}\leq 1,
  \qquad
  &h=\frac{\sqrt{D_N}}{6M\sqrt{n}\,(N+1)^{3/4}}, \qquad N+1\ge n^{2}  
  \label{eq:dyn_tuning} \\
\begin{split}
\text{yields} \qquad  \E\big[\DRegret_N\big]
  &\le\big(\sqrt n(2\bar L_0 + 3M) + 6M\sqrt{n D_N}\big)(N + 1)^{3/4} \\
  &=O\Big(M\big(n+n^{3/4}\sqrt{\PN}\big)(N+1)^{\tfrac{3}{4}}\Big). 
\end{split} \label{eq:dyn_rate}
\end{align}

\noindent (ii) If $|f_k(S)-f_{k^+}(S)|\le V$ for all $S\subset[n]$
and for all $k\in \{0,\ldots,N\}$, then for any $N\ge0$,
\begin{equation}\label{eq:dyn_general_V}
  \frac{1}{N+1}\sum_{k=0}^{N}
  \Big(\E\big[f^L_k(x_k)\big]-f^L_k(x^\ast_k)\Big)
  \!\le\!
  \frac{D_N}{2h(N+1)}
  \!+\! \mu\bar L_0
  \!+\! \frac{hn^2(1+2\mu)^2}{\mu^2}\Big(V^2+\frac{4M^2\mu\sqrt n}{1+2\mu}\Big).
\end{equation}
Choosing $h$, $\mu$ and $N$ such that
\begin{align}
  \mu=\frac{n}{(N+1)^{1/3}}  \leq 1, 
  \ \ 
  h=&\frac{\sqrt{D_N}}{4\sqrt3\,Mn^{3/4}(N+1)^{2/3}},  \ \  N+1\ge n^{3},  \ \ V\le \sqrt{\tfrac{4}{3}}M n^{1/4} \sqrt{\mu} \label{eq:dyn_tuning_V} \\
\begin{split}
\text{yields} \ \  \qquad  \E\big[\DRegret_N\big]
  &\le \big(n(2\bar L_0+3M)+4\sqrt3Mn^{3/4}\sqrt{D_N}\big)(N+1)^{\tfrac{2}{3}} \\
  &= O\big((n\bar L_0+Mn^{5/4}+Mn\sqrt{\PN})(N+1)^{2/3}\big).
\end{split} \label{eq:dyn_rate_V}
\end{align}
\end{theorem}

 {Theorems~\ref{thm:static} and~\ref{thm:dynamic} count only the
exploration play in the regret. If the base play is counted as well, the
$O(N^{2/3})$ rates are preserved provided $V=O(N^{-1/3})$
(Remark~\ref{rem:charge_both} in Appendix~\ref{app:rem}).}
 {The tuned step sizes require an upper bound on $\PN$, as is standard
in dynamic regret analyses \citep{zinkevich2003online}, whereas
\eqref{eq:dyn_general} and \eqref{eq:dyn_general_V} hold for any
$h,\mu>0$.}

\section{Numerical Example}
\label{sec:numerics}
In this section, we present a numerical example to illustrate the theoretical findings of Section~\ref{sec:main}.
The experiment is implemented in Python and run on a laptop with an Intel Core
i7-1265U processor.\footnote{The code is available at \url{https://github.com/amirali78frz/ZO_bandit_submodular.git}.}
%
%
We consider the time-varying submodular functions
\begin{equation}\label{eq:num_family}
f_k(S)=\alpha\sqrt{\textstyle\sum_{i\in S}w_i}
-\textstyle\sum_{i\in S}b_k(i),\qquad S\subset[n],
\end{equation}
where $w_i>0$ and
$
b_k(i)=\bar b(i)+A\sin(2\pi k/\Pi+\phi_i).
$
 The first term is a concave, non-decreasing
function of a non-negative modular function and is therefore submodular. The
second term is modular, so $f_k$ is submodular, and $f_k(\emptyset)=0$. The two
terms act in opposition. The time-varying
rewards induce a moving minimiser, with path length controlled by $A$ and
$\Pi$.
Unless stated otherwise, we use $n=10$, $\alpha=3$, $A=1$, $\Pi=500$, and
 a horizon of $75000$ plays. The vectors $w$, $\bar b$ and $\phi$ are drawn once
uniformly at random. Since $n$ is small, all $2^n$ subsets can be enumerated,
allowing both regret and minimiser path length to be computed exactly. 
 {The} drift between consecutive plays is $V\approx 0.06$, so the drift condition of item~(ii) of Theorems~\ref{thm:static} and~\ref{thm:dynamic} holds.

\begin{figure}[t]
  \centering
  \includegraphics[width=0.45\textwidth]{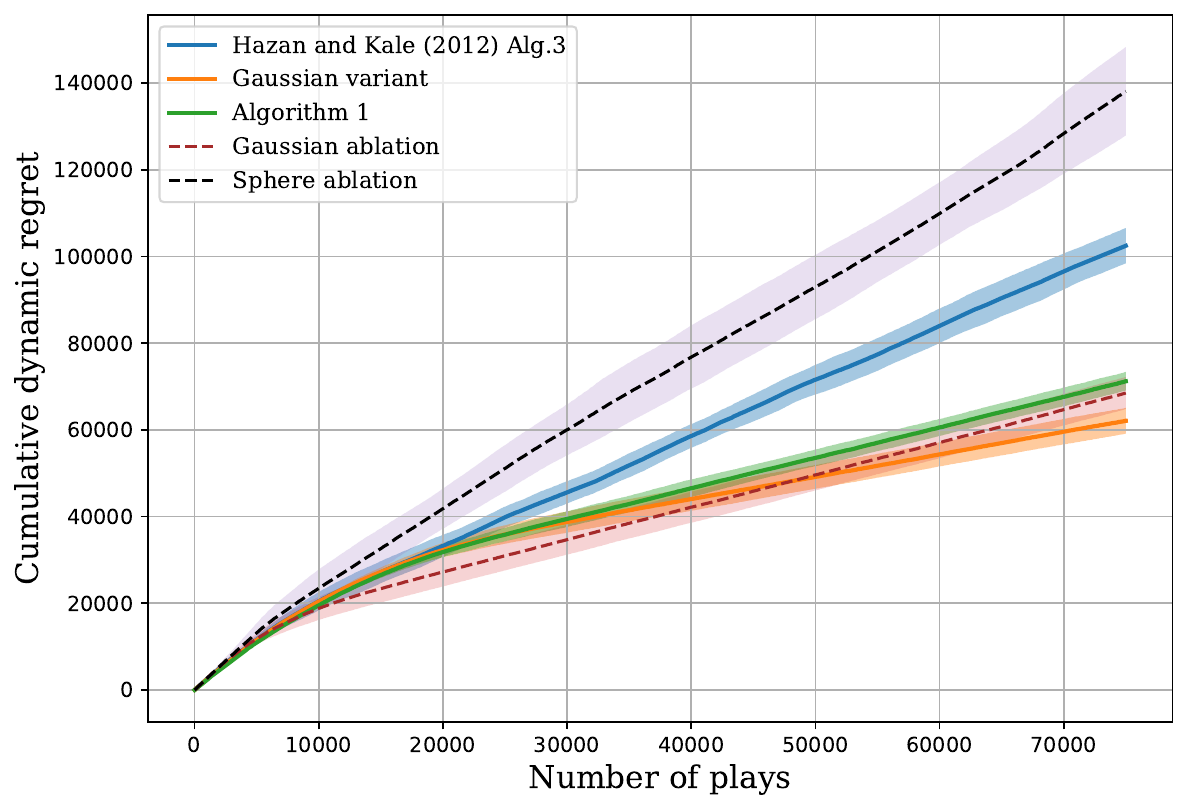}
  \includegraphics[width=0.45\textwidth]{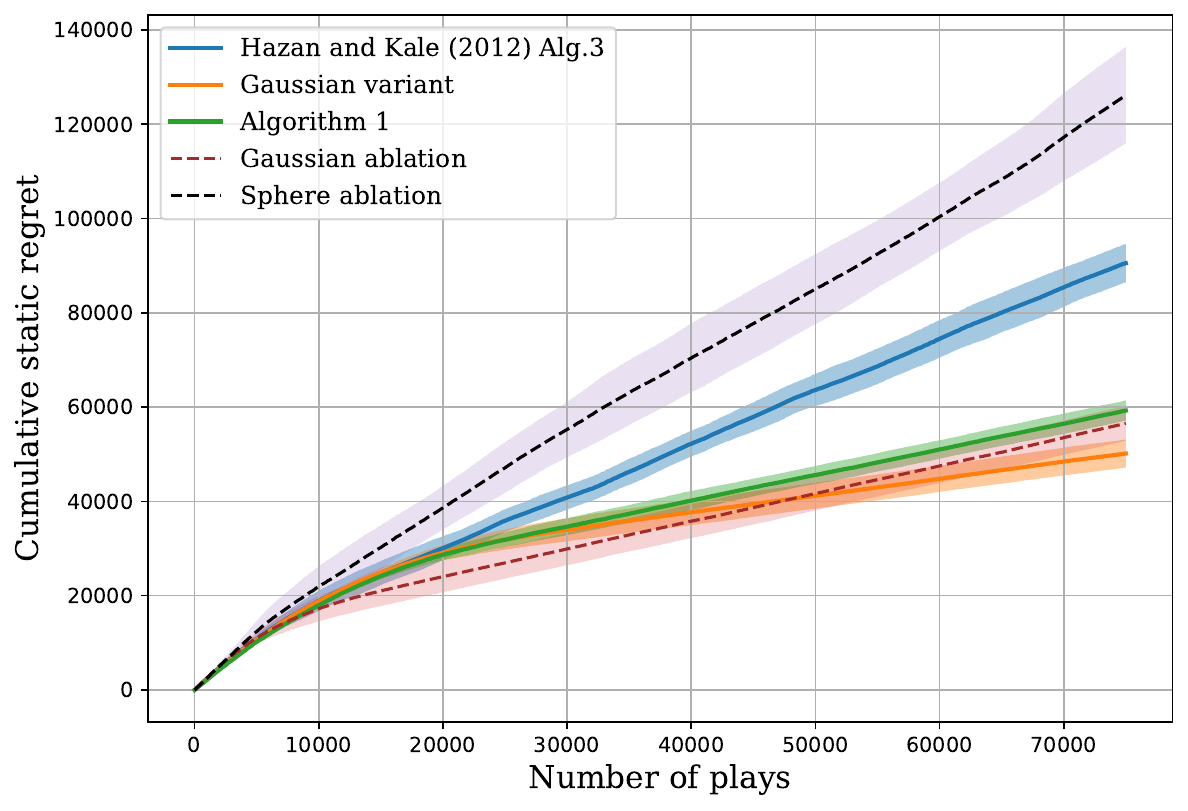}
  \caption{Cumulative dynamic regret (left) and static regret
  (right) against the number of plays.}
  \label{fig:regret}
\end{figure}

\begin{figure}[t]
  \centering
  \includegraphics[width=0.45\textwidth]{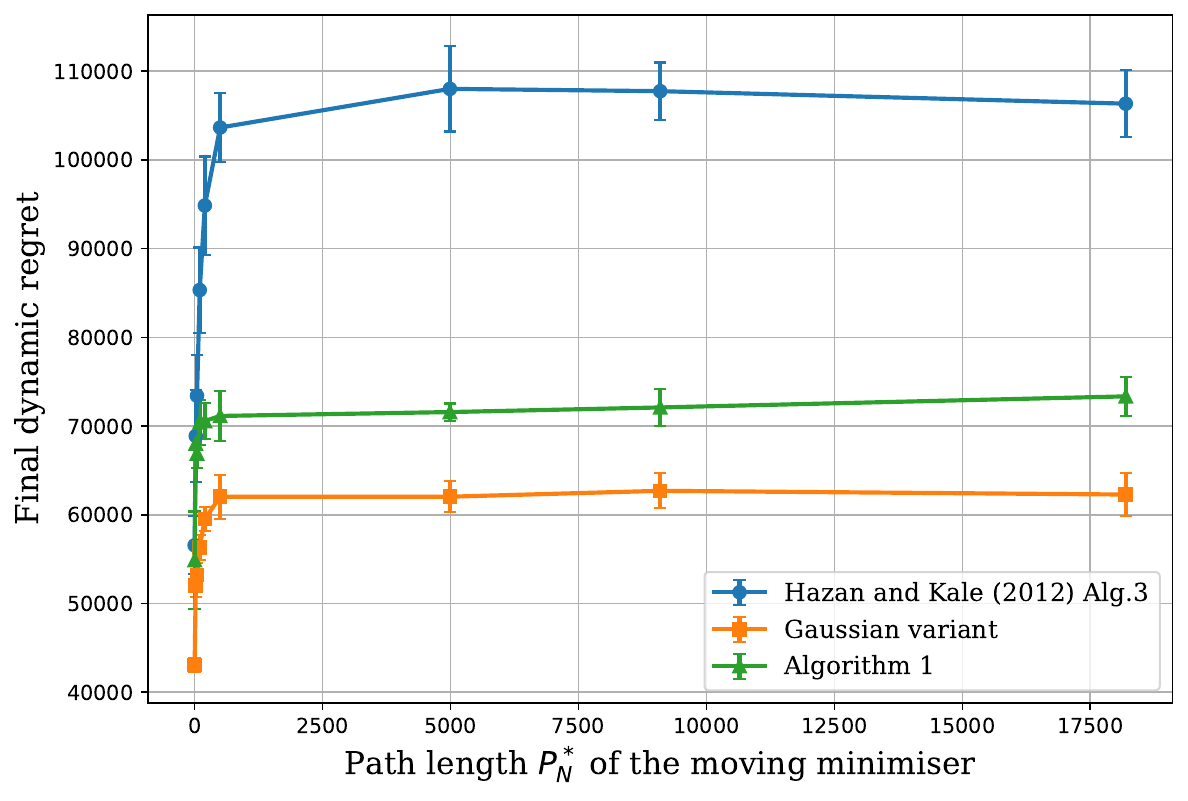}
  \includegraphics[width=0.45\textwidth]{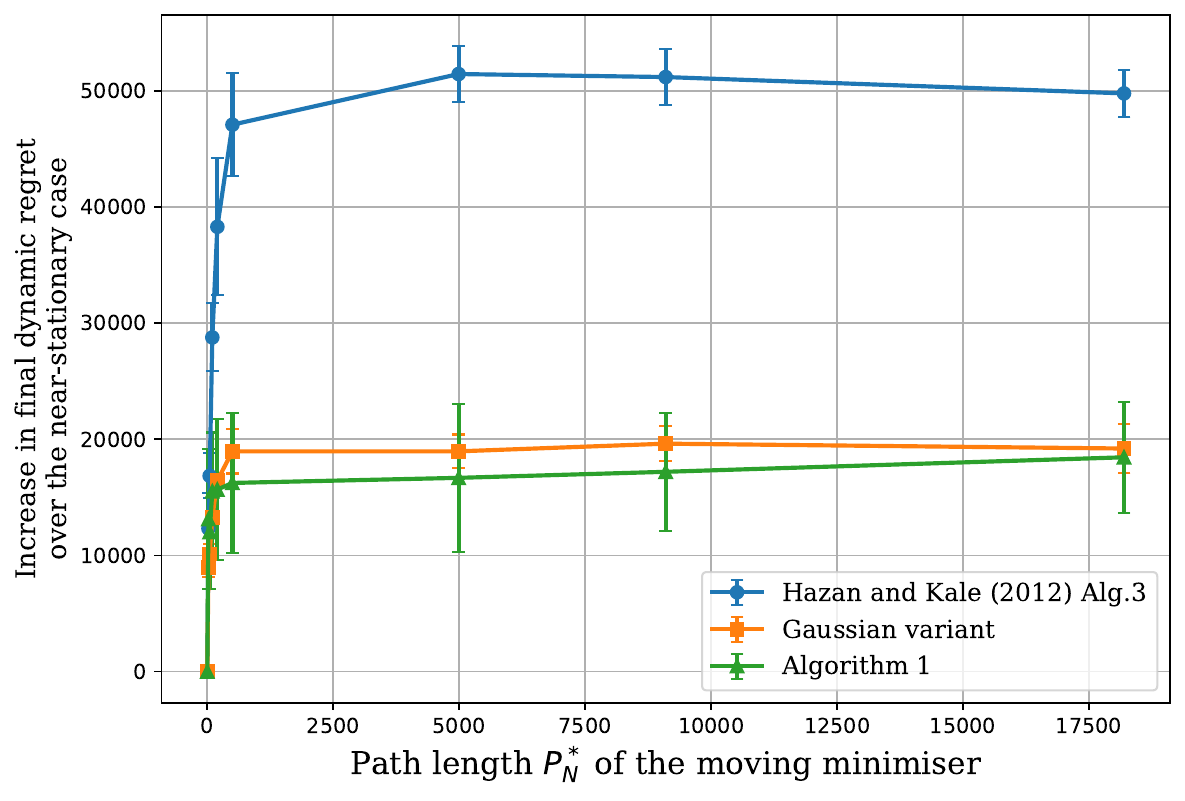}
\caption{Final dynamic regret vs.\ $\PN$ (left), and its increase over the near-stationary case (right).}
  \label{fig:pathlength}
\end{figure}

We compare Algorithm~\ref{alg:main} with Algorithm~3 of
\citet{hazan2012online} (the baseline), as well as Algorithm~\ref{alg:main}'s Gaussian counterpart described in
Remark~\ref{rem:gaussian_fail}. The Gaussian variant draws $u_k\sim\mathcal N(0,I)$ and $\tau_k\sim U[0,1]$,
plays $S_{\tau_k}(x_k+\mu u_k)$ and $S_{\tau_k}(x_k)$, and uses
$\hat g_k=\tfrac{1}{\mu}(c_k-c_{k^+})u_k$. For each two-point method, we also consider the one-point ablation of
Remark~\ref{rem:onepoint}, referred to as the sphere and Gaussian
ablations. Every play incurs its
observed cost and is  {counted in the regret}. Hence two-point methods perform half
as many updates as one-point methods over the same horizon. This
counting is stricter than the convention of Section~\ref{subsec:static}, under
which only the exploration play is counted in the regret. The regrets plotted
in Figures~\ref{fig:regret} and~\ref{fig:pathlength} include the cost of every play (Remark~\ref{rem:charge_both}).

Each configuration is run with five independent seeds on the same instance.
Results report the mean with one-standard-deviation bands. Parameters are
tuned for each method by minimising mean final dynamic regret over
$\eta,h\in\{10^{-5},10^{-4},10^{-3}\},$ 
$\delta\in\{0.05,0.15,0.5,0.6,0.75\},$ 
$\mu\in\{10^{-3},10^{-2},10^{-1},0.25,0.5\}.$
The selected parameters are $(\eta,\delta)=(10^{-4},0.15)$ for the baseline,
$(h,\mu)=(10^{-4},0.1)$ for Algorithm~\ref{alg:main}, the Gaussian variant
and the Gaussian ablation, and $(h,\mu)=(10^{-4},0.25)$ for the sphere
ablation.
For Figure~\ref{fig:pathlength},
this tuning is repeated for each drift period.

\begin{figure}[t]
  \centering
  \includegraphics[width=0.9\textwidth]{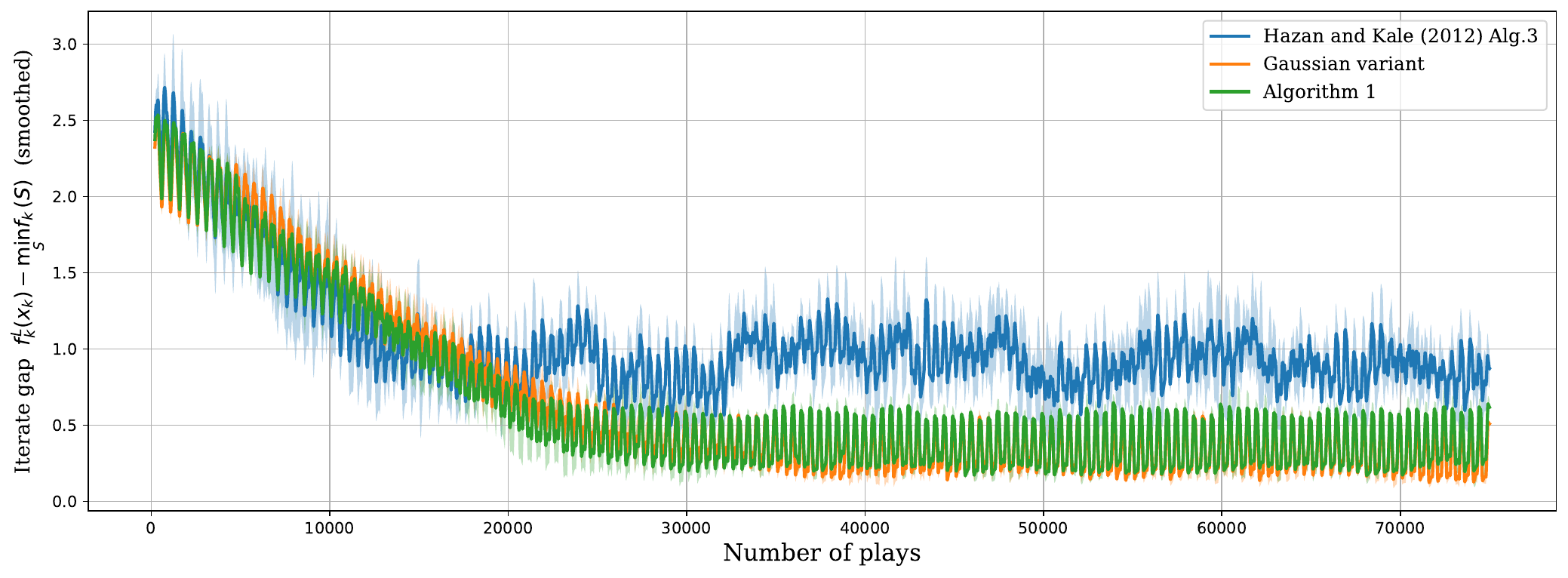}
  \caption{Iterate optimality gap $f^L_k(x_k)-\min_S f_k(S)$ against the number
  of plays.}
  \label{fig:convergence}
\end{figure}

Figure~\ref{fig:regret} compares static and dynamic regret for an instance with \(\PN=997.5\). Both two-point methods outperform the baseline, with final dynamic regret of \(62{,}093\pm2{,}975\) for the Gaussian variant, \(71{,}209\pm2{,}162\) for Algorithm~\ref{alg:main}, and \(102{,}466\pm4{,}068\) for the baseline. The Gaussian and sphere ablations reach $68{,}502\pm3{,}707$ and $138{,}087\pm10{,}205$, respectively, highlighting the importance of the base evaluation for Algorithm~\ref{alg:main}.
Averaged over the second half of the horizon, the optimality gaps in Figure~\ref{fig:convergence} are $0.325$, $0.368$ and $0.920$ for the Gaussian variant, Algorithm~\ref{alg:main} and the baseline, respectively. 
Figure~\ref{fig:pathlength} shows that Algorithm~\ref{alg:main} and the baseline perform similarly for $\PN\approx0$, whereas the baseline deteriorates much faster as $\PN$ grows. The advantage of Algorithm~\ref{alg:main} thus stems mainly from tracking the moving minimiser.

\section{Conclusions and Future Work}
\label{sec:conclusion}

In this paper, we studied bandit online submodular minimisation and proposed a zeroth-order oracle based on the threshold representation of the Lovász extension. Unit ball smoothing enables a direction-independent threshold range, making the base evaluation a zero-mean control variate and yielding an unbiased estimator of the smoothed gradient. We established \(O(N^{3/4})\) static regret for bounded costs, improved to \(O(N^{2/3})\) under sufficiently small drift. Moreover, we establish dynamic regret bounds of order $O\big((1+\sqrt{P^\ast_N})N^{3/4}\big)$ in general and $O\big((1+\sqrt{P^\ast_N})N^{2/3}\big)$ under a sufficiently small drift,
where $P^\ast_N$ is the path length of the minimiser sequence. In experiments, the method attains lower regret than the algorithm of \citet{hazan2012online}, with the advantage growing with the path length, and the base evaluation proves essential.

Several limitations suggest directions for future work. First, whether a matching $\Omega(N^{2/3})$ lower bound on the static regret holds for bandit online SFM remains open. Further directions include extending the analysis to the Gaussian variant of Section~\ref{sec:numerics}, which attains the lowest regret in our experiments but is not covered by our analysis. Future work could extend the analysis to approximately submodular functions, min-max problems, and distributed or privacy-preserving settings.

\bibliography{L4DC2027}

\appendix
\section{Proof of the main results}\label{app:proof}
In this section, we provide the proofs of the main Lemmas and Theorems given in Section~\ref{sec:main}.

\smallskip

\begin{proofof}\textbf{Lemma~\ref{lem:unbiased}:}
Write $\hat g_\mu(x)=\tfrac{n(b-a)}{\mu}(f_{k}(S_{\tau}(x+\mu u))- f_{k^+}(S_{\tau}(x)))u,$ and split the
expectation into 
two terms.
The quantity $ f_{k^+}(S_\tau(x))$
depends on $\tau$ and the fixed point $x$, but not on $u$. Since $u\sim\mathbb S$
is independent of $\tau$ and $\E_u[u]=0$,
\begin{equation}\label{eq:pf_cv}
  \E_{u,\tau}\big[ f_{k^+}(S_\tau(x))\,u\big]
  = \E_\tau\big[ f_{k^+}(S_\tau(x))\big]\,\E_u[u] = 0 .
\end{equation}
 {Accordingly,} 
the base evaluation contributes nothing to the mean  {and it} 
is a zero-mean
control variate.
 {To proceed, we condition the last expression on} 
$u$ and integrate over $\tau$.  {Then, by}
Remark~\ref{rem:extended}, and since $[a,b]$ brackets $x+\mu u$ for every
$u\in\mathbb S$,  {it holds that}
\begin{equation}\label{eq:pf_choquet}
  (b-a)\,\E_\tau\big[ f_k(S_\tau(x+\mu u))\big]
  =  f^L_k(x+\mu u) - a\, f_k([n]).
\end{equation}
Substituting \eqref{eq:pf_cv} and \eqref{eq:pf_choquet} into the expectation of
\eqref{eq:oracle},
\begin{align}
  \E_{u,\tau}\big[\hat g_\mu(x)\big]
  &= \frac{n}{\mu}\,\E_u\Big[\big( f^L_k(x+\mu u) - a\, f_k([n])\big)u\Big]
  \nonumber\\
  &= \frac{n}{\mu}\,\E_u\big[ f^L_k(x+\mu u)\,u\big]
   - \frac{n}{\mu}\,a\, f_k([n])\,\E_u[u]
  = \nabla  f^L_{\mu,k}(x),\label{eq:pf_final}
\end{align}
where the constant term vanishes by $\E_u[u]=0$ and the last equality is the
sphere gradient identity \eqref{eq:sphere_grad} applied to $F=f^L_k$. This
proves \eqref{eq:unbiased}.

For the second-moment bounds  {\eqref{eq:unbiased} and \eqref{eq:secondmoment_V},} since $\norm{u}=1$,  {we first observe that} 
\begin{equation}\label{eq:pf_norm}
  \norm{\hat g_\mu(x)}^2
  = \frac{n^2(b-a)^2}{\mu^2}\,(c_k - c_{k^+})^2,
  \qquad c_k = f_k(S_\tau(x+\mu u)),\ \ c_{k^+}=f_{k^+}(S_\tau(x)).
\end{equation}

 {Then, each} 
$c_k$ and
$c_{k^+}$ is bounded by $M$, so $|c_k-c_{k^+}|\le 2M$ giving \eqref{eq:unbiased}.

 {To prove \eqref{eq:secondmoment_V} we introduce}
the intermediate value $f_k(S_\tau(x))$ ({with the} same threshold) and split
\begin{equation}\label{eq:pf_split}
  c_k - c_{k^+}
  = \underbrace{\big(f_k(S_\tau(x+\mu u))-f_k(S_\tau(x))\big)}_{=:\,\Delta_\tau(u)}
  \;+\; \underbrace{\big(f_k(S_\tau(x))-f_{k^+}(S_\tau(x))\big)}_{=:\,\delta_\tau},
\end{equation}
so that $(c_k-c_{k^+})^2 \le 2\Delta_\tau(u)^2 + 2\delta_\tau^2$. 
The drift term
obeys $|\delta_\tau|\le V$ by 
 {assumption and} hence $\delta_\tau^2\le V^2$.

For $\Delta_\tau(u)$, note first that if the two level sets coincide then
$\Delta_\tau(u)=0$. 
Hence
\begin{equation}\label{eq:pf_incl}
  \big\{\Delta_\tau(u)\neq 0\big\}
  \;\subset\;
  \big\{S_\tau(x+\mu u)\neq S_\tau(x)\big\},
\end{equation}
 {It therefore suffices to bound the probability of the event on the
right-hand side.}  {Coordinate} $i$ belongs to exactly one of the two
level sets if and only if $\tau$ lies between $x(i)$ and $x(i)+\mu u(i)$.
 {This interval has length $\mu|u(i)|$ and is contained in $[a,b]$ for
every $u\in\mathbb S$. Since $\tau\sim U[a,b]$, the event therefore has
probability $\mu|u(i)|/(b-a)$. Moreover, the two level sets differ if and
only if at least one coordinate belongs to exactly one of them. Hence,} by
\eqref{eq:pf_incl} and a union bound over $i\in[n]$,
\begin{equation}\label{eq:pf_cross}
  \Pr\nolimits_\tau\big[\Delta_\tau(u)\neq 0\big]
  \;\le\; \Pr\nolimits_\tau\big[S_\tau(x+\mu u)\neq S_\tau(x)\big]
  \;\le\; \sum_{i=1}^n \frac{\mu|u(i)|}{b-a}
  \;=\; \frac{\mu\,\norm{u}_1}{b-a}.
\end{equation}
Whenever $\Delta_\tau(u)\neq 0$ we have $|\Delta_\tau(u)|\le 2M$. 
Hence, conditioning on $u$ and taking the expectation
over $\tau$,  {leads to the estimates}
\begin{equation}\label{eq:pf_Dbound}
  \E_\tau\big[\Delta_\tau(u)^2\big]
  \;\le\; (2M)^2\,\Pr\nolimits_\tau\big[\Delta_\tau(u)\neq 0\big]
  \;\le\; \frac{4M^2\mu\,\norm{u}_1}{b-a}.
\end{equation}
Taking the expectation over $u$ and using
$\E_u[\norm{u}_1]\leq\sqrt n$   {provides the inequality}
\begin{equation}\label{eq:pf_EDbound}
  \E_{u,\tau}\big[\Delta_\tau(u)^2\big]
  \;\le\; \frac{4M^2\mu\sqrt n}{b-a}.
\end{equation}
Combining $(c_k-c_{k^+})^2\le 2\Delta_\tau(u)^2+2V^2$ with \eqref{eq:pf_norm},
taking expectations and substituting \eqref{eq:pf_EDbound},  {guarantees that the expectation satisfies}
\begin{align}
  \E_{u,\tau}\big[\norm{\hat g_\mu(x)}^2\big]
  &\le\! \frac{n^2(b-a)^2}{\mu^2}
     \Big( 2\E_{u,\tau}[\Delta_\tau(u)^2] \!+\! 2V^2 \Big) 
  \!\le\! \frac{n^2(b-a)^2}{\mu^2}
     \Big( \frac{8M^2\mu\sqrt n}{b-a} \!+\! 2V^2 \Big),
\end{align}
which is \eqref{eq:secondmoment_V} since $b-a=1+2\mu$,  {and thus completes the proof}.
\end{proofof}

\smallskip

\begin{proofof}\textbf{Theorem~\ref{thm:static}:}
Let $r_k=\norm{x_k-x^\ast}$. By non-expansiveness of the projection onto the
convex set $[0,1]^n$ and the update $x_{k+1}=\Proj(x_k-h\hat g_k)$,  {we have}
\begin{equation}\label{eq:pf_st_nonexp}
  r_{k+1}^2
  \le \norm{x_k-x^\ast-h\hat g_k}^2
  = r_k^2 - 2h\inner{\hat g_k}{x_k-x^\ast} + h^2\norm{\hat g_k}^2 .
\end{equation}
For item (i) taking the expectation of \eqref{eq:pf_st_nonexp} with respect to $u_k,\tau_k$ and invoking Lemma~\ref{lem:unbiased} give
\begin{equation}\label{eq:pf_st_cond}
  \E\big[r_{k+1}^2\big]
  \le r_k^2 - 2h\inner{\nabla f^L_{\mu,k}(x_k)}{x_k-x^\ast}
  + \frac{4h^2 M^2 n^2(1+2\mu)^2}{\mu^2}.
\end{equation}
Since $f_k$ is submodular, $f^L_k$ is convex (Lemma~\ref{lem:lovasz_props}), and
hence $f^L_{\mu,k}$ is convex. Convexity yields
$\langle{\nabla f^L_{\mu,k}(x_k)},{x_k-x^\ast}\rangle\ge f^L_{\mu,k}(x_k)-f^L_{\mu,k}(x^\ast)$.
Substituting  {this estimate} into \eqref{eq:pf_st_cond} and rearranging  {yields}
\begin{equation}\label{eq:pf_st_smoothed}
  f^L_{\mu,k}(x_k)-f^L_{\mu,k}(x^\ast)
  \le \frac{r_k^2-\E[r_{k+1}^2]}{2h}
  + \frac{2h M^2 n^2(1+2\mu)^2}{\mu^2}.
\end{equation}
Considering the bias bound $0\le f^L_{\mu,k}(x)-f^L_k(x)\le\mu \bar L_0$, where
the lower bound follows from the convexity of $f^L_k$, and the upper bound from $\bar L_0$-Lipschitzness with
$\norm{v}\le 1$, with the lower bound applied at $x_k$ and the upper bound
applied at $x^\ast$, we have
\begin{equation}\label{eq:pf_st_perstep}
  f^L_k(x_k)-f^L_k(x^\ast)
  \le \frac{r_k^2-\E[r_{k+1}^2]}{2h}
  + \mu \bar L_0
  + \frac{2h M^2 n^2(1+2\mu)^2}{\mu^2}.
\end{equation}
Taking the expectation over the randomness up to round $k-1$, summing
\eqref{eq:pf_st_perstep} from $k=0$ to $N$,  {using a} telescoping  {argument} with
$r_0^2\le n$ (as $x_0,x^\ast\in[0,1]^n$) and $\E[r_{N+1}^2]\ge 0$, and dividing by
$N+1$ gives \eqref{eq:static_general}.

It remains to relate \eqref{eq:static_general} to the incurred cost.
Considering Lemma~\ref{lem:lovasz_props}(iii) applied to the submodular
function $\sum_{k=0}^N f_k$, with Lov\'asz extension
$\sum_{k=0}^N f^L_k$, we have
$\sum_{k=0}^N f_k(S^\ast)=\sum_{k=0}^N f^L_k(x^\ast)$, where $S^\ast$ is the minimiser of
$\sum_{k=0}^N f_k(S)$.
Considering the projection step in
Algorithm~\ref{alg:main}, we know that $x_k\in[0,1]^n$ for all
$k\in\{0,\dots,N\}$. 
Considering Remark~\ref{rem:extended}, since the fixed range
$[a,b]=[-\mu,1+\mu]$ brackets $x_k+\mu u_k$ for every $u_k\in\mathbb{S}$, with
$a=-\mu$ and $b-a=1+2\mu$, we
have $\E_{\tau_k}[f_k(S_{\tau_k}(x_k+\mu u_k))]=\frac{f^L_k(x_k+\mu u_k)+\mu f_k([n])}{1+2\mu}$, and by
$\bar L_0$-Lipschitzness with $\norm{u_k}=1$,
\begin{equation}\label{eq:pf_st_explore}
  \E_{\tau_k}\big[f_k(S_{\tau_k}(x_k+\mu u_k))\big]
  = \frac{f^L_k(x_k+\mu u_k)+\mu f_k([n])}{1+2\mu}
  \le \frac{f^L_k(x_k)+\mu\bar L_0+\mu f_k([n])}{1+2\mu} .
\end{equation}
Moreover, $\tau_k$ and $u_k$ are independent. Considering these facts, taking
the expectation of $f_k(S_k)-f_k(S^\ast)$ with respect to $\tau_k$,
taking the expectation with respect to the randomness up to round $k-1$,
summing from $k=0$ to $N$, and using \eqref{eq:static_general}, whose
right-hand side is nonnegative, together with $1+2\mu\ge 1$,
$|f_k([n])|\le M$ and $|f^L_k(x^\ast)|\le M$ (by Remark~\ref{rem:lovae}, as $x^\ast\in[0,1]^n$),
we obtain
\begin{align}
  \frac{1}{ N+1}\E[\Regret_N]
  &\le\frac{1}{ N+1} \sum_{k=0}^{N}\Big(\frac{\big(\E[f^L_k(x_k)]-f^L_k(x^\ast)\big)+\mu\bar L_0}{1+2\mu}
  +\frac{\mu\big(f_k([n])-2f^L_k(x^\ast)\big)}{1+2\mu}\Big)
  \nonumber\\
  &\le\frac{n}{2h(N+1)}+2\mu\bar L_0
      +\frac{2hM^2n^2(1+2\mu)^2}{\mu^2}+3\mu M.
  \label{eq:pf_st_total}
\end{align}
With the choices \eqref{eq:static_tuning} and $(1+2\mu)\le 3$, we obtain
\eqref{eq:static_rate}, which proves item~(i).

For item~(ii) the argument is identical, with the second-moment bound
\eqref{eq:unbiased} replaced by \eqref{eq:secondmoment_V} in
\eqref{eq:pf_st_cond}. Carrying
$\tfrac{h}{2}\E\norm{\hat g_k}^2
\le \tfrac{hn^2(1+2\mu)^2}{\mu^2}\big(V^2+\tfrac{4M^2\mu\sqrt n}{1+2\mu}\big)$
through \eqref{eq:pf_st_smoothed}--\eqref{eq:pf_st_total} yields
\eqref{eq:static_general_V} and
\begin{equation}\label{eq:pf_st_total_V}
  \frac{1}{ N+1}\E[\Regret_N]
  \;\le\; \frac{n}{2h(N+1)}+2\mu\bar L_0
  +\frac{(1+2\mu)^2hn^2V^2}{\mu^2}+\frac{12hM^2n^{5/2}}{\mu}
  +3\mu M.
\end{equation}
With the choices \eqref{eq:static_tuning_V} and $(1+2\mu)\le 3$, we obtain
\eqref{eq:static_rate_V}, which proves item~(ii).
\end{proofof}

\smallskip

\begin{proofof}\textbf{Theorem~\ref{thm:dynamic}:}
Let $e_k=\norm{x_k-x^\ast_k}$. Then
\begin{equation}\label{eq:pf_dy_split}
  e_{k+1}=\norm{x_{k+1}-x^\ast_{k+1}+x^\ast_k-x^\ast_k}
  \le\norm{x_{k+1}-x^\ast_k}+\norm{x^\ast_{k+1}-x^\ast_k},
\end{equation}
and since $x_{k+1},x^\ast_k\in[0,1]^n$ we have $\norm{x_{k+1}-x^\ast_k}\le\sqrt n$,
whence
\begin{equation}\label{eq:pf_dy_sq}
  e_{k+1}^2\le\norm{x_{k+1}-x^\ast_k}^2+\norm{x^\ast_{k+1}-x^\ast_k}^2
  +2\sqrt n\,\norm{x^\ast_{k+1}-x^\ast_k}.
\end{equation}
By non-expansiveness of the projection onto $[0,1]^n$ and the update
$x_{k+1}=\Proj(x_k-h\hat g_k)$,
\begin{equation}\label{eq:pf_dy_nonexp}
  \norm{x_{k+1}-x^\ast_k}^2\le\norm{x_k-x^\ast_k-h\hat g_k}^2
  =e_k^2-2h\inner{\hat g_k}{x_k-x^\ast_k}+h^2\norm{\hat g_k}^2 .
\end{equation}
Substituting \eqref{eq:pf_dy_nonexp} into \eqref{eq:pf_dy_sq} and, for item~(i),
taking the expectation with respect to $u_k,\tau_k$ and invoking
Lemma~\ref{lem:unbiased} give
\begin{equation}\label{eq:pf_dy_cond}
  \E\big[e_{k+1}^2\big]\!\le\! e_k^2-\!2h\inner{\nabla f^L_{\mu,k}(x_k)}{x_k-x^\ast_k}
  \!+\frac{4h^2M^2n^2(1\!+\!2\mu)^2}{\mu^2}
  +\norm{x^\ast_{k+1}\!-\!x^\ast_k}^2\!+\!2\sqrt n\norm{x^\ast_{k+1}\!-\!x^\ast_k}.
\end{equation}
Since $f_k$ is submodular, $f^L_k$ is convex (Lemma~\ref{lem:lovasz_props}), and
hence $f^L_{\mu,k}$ is convex. Convexity yields
$\langle{\nabla f^L_{\mu,k}(x_k)},{x_k-x^\ast_k}\rangle\ge
f^L_{\mu,k}(x_k)-f^L_{\mu,k}(x^\ast_k)$. Moreover
$\norm{x^\ast_{k+1}-x^\ast_k}^2\le\sqrt n\norm{x^\ast_{k+1}-x^\ast_k}$, since
$x^\ast_k,x^\ast_{k+1}\in\{0,1\}^n$. Substituting both into \eqref{eq:pf_dy_cond} and
rearranging, leads to
\begin{equation}\label{eq:pf_dy_smoothed}
  f^L_{\mu,k}(x_k)-f^L_{\mu,k}(x^\ast_k)
  \le\frac{e_k^2-\E[e_{k+1}^2]}{2h}
  +\frac{2hM^2n^2(1+2\mu)^2}{\mu^2}
  +\frac{3\sqrt n}{2h}\norm{x^\ast_{k+1}-x^\ast_k}.
\end{equation}
Next, we pass from the smoothed extension $f^L_{\mu,k}$ to $f^L_k$ via the bias bound
$0\le f^L_{\mu,k}(x)-f^L_k(x)\le\mu\bar L_0$. Here the lower bound follows
from the convexity of $f^L_k$, and
the upper bound from $\bar L_0$-Lipschitzness with $\norm{v}\le 1$. With the
lower bound applied at $x_k$ and the upper bound applied at $x^\ast_k$, it holds that
\begin{equation}\label{eq:pf_dy_perstep}
  f^L_k(x_k)-f^L_k(x^\ast_k)
  \le\frac{e_k^2-\E[e_{k+1}^2]}{2h}+\mu\bar L_0
  +\frac{2hM^2n^2(1+2\mu)^2}{\mu^2}
  +\frac{3\sqrt n}{2h}\norm{x^\ast_{k+1}-x^\ast_k}.
\end{equation}
Finally, taking the expectation over the randomness up to round $k-1$, summing
\eqref{eq:pf_dy_perstep} from $k=0$ to $N$ (and setting $x^\ast_{N+1}\defe x^\ast_N$, so that the last term of the sum vanishes), telescoping with $e_0^2\le n$ and
$\E[e_{N+1}^2]\ge0$, using
$\sum_{k=0}^{N}\norm{x^\ast_{k+1}-x^\ast_k}\le\PN$, and dividing by
$N+1$ gives \eqref{eq:dyn_general}.

It remains to relate \eqref{eq:dyn_general} to the incurred cost. By
Lemma~\ref{lem:lovasz_props}(iii), $f_k(S^\ast_k)=f^L_k(x^\ast_k)$ for every $k$.
By the projection step in Algorithm~\ref{alg:main}, $x_k\in[0,1]^n$, and
since $\norm{u_k}=1$, the fixed range $[-\mu,1+\mu]$ contains every
coordinate of $x_k+\mu u_k$. Moreover, $\tau_k$ is independent of $u_k$.
Hence, conditioning on $u_k$, \eqref{eq:pf_st_explore} applies and gives
\begin{equation}\label{eq:pf_dy_cost}
  \E_{\tau_k}\big[f_k(S_k)\big]-f_k(S^\ast_k)
  \le \frac{f^L_k(x_k)-f^L_k(x^\ast_k)+\mu\bar L_0}{1+2\mu}
  +\frac{\mu\big(f_k([n])-2f^L_k(x^\ast_k)\big)}{1+2\mu}.
\end{equation}
The last term is at most $3\mu M$, since $|f_k([n])|\le M$ and
$|f^L_k(x^\ast_k)|=|f_k(S^\ast_k)|\le M$. We now take the full expectation of
\eqref{eq:pf_dy_cost} and sum from $k=0$ to $N$. The right-hand side of
\eqref{eq:dyn_general} is nonnegative and $1+2\mu\ge1$, so the sum of the
first terms is at most $N+1$ times the right-hand side of
\eqref{eq:dyn_general} plus $\mu\bar L_0$. Dividing by $N+1$, we obtain
\begin{equation}\label{eq:pf_dy_total}
  \frac{1}{N+1}\E\big[\DRegret_N\big]
  \le\frac{D_N}{2h(N+1)}+2\mu\bar L_0
  +\frac{2hM^2n^2(1+2\mu)^2}{\mu^2}+3\mu M.
\end{equation}
With the choices \eqref{eq:dyn_tuning} and $(1+2\mu)\le3$, we obtain
\eqref{eq:dyn_rate}, which proves item~(i).

For item~(ii) the argument is identical, with the second-moment bound
\eqref{eq:unbiased} replaced by \eqref{eq:secondmoment_V} in
\eqref{eq:pf_dy_cond}. Carrying
$\tfrac h2\E\norm{\hat g_k}^2\le
\tfrac{hn^2(1+2\mu)^2}{\mu^2}\big(V^2+\tfrac{4M^2\mu\sqrt n}{1+2\mu}\big)$
through \eqref{eq:pf_dy_smoothed}--\eqref{eq:pf_dy_total} yields
\eqref{eq:dyn_general_V}, and with the choices \eqref{eq:dyn_tuning_V}, 
the stated drift condition and $(1+2\mu)\le3$ we obtain \eqref{eq:dyn_rate_V}.
\end{proofof}

\section{Additional remarks}\label{app:rem}

In this section, we present additional remarks for better explanation of the given results.

\begin{remark}\label{rem:gaussian_fail}
Unbiasedness in \eqref{eq:unbiased} relies on $\tau$ being independent of
$u$. This independence holds for \eqref{eq:oracle} because $\norm{u}=1$,
so the fixed range $[a,b]=[-\mu,1+\mu]$ contains every coordinate of
$x+\mu u$ for all $u\in\mathbb S$. The following two alternatives fail.
 
\emph{(i) A $u$-dependent range.} Suppose the range is chosen as the
smallest interval containing $[0,1]$ and all coordinates of $x+\mu u$,
i.e.\ $a(u)=\min\{0,\min_i(x+\mu u)(i)\}$ and
$b(u)=\max\{1,\max_i(x+\mu u)(i)\}$. Then $\tau$ depends on $u$. Applying
\eqref{eq:lovasz_boxed} to both plays, their constant terms combine to
$-a(u)\big(f_k([n])-f_{k^+}([n])\big)$. This term is correlated with $u$,
so the mean of the oracle acquires the bias
$-\tfrac{n}{\mu}\big(f_k([n])-f_{k^+}([n])\big)\E_u[a(u)u]$, which is
nonzero in general.

\emph{(ii) Gaussian smoothing.} If $u\sim\mathcal N(0,I)$, then $u$ is
unbounded and no fixed finite range contains $x+\mu u$ for all $u$. Drawing
$\tau$ from $[0,1]$ instead is equivalent to rounding the clipped point
$\Proj(x+\mu u)$, so the oracle targets the smoothing of
$f^L_k(\Proj(\cdot))$, which need not be convex on $\R^n$. For example, for
$n=1$ and $f_k(\{1\})=c>0$, the map $y\mapsto f^L_k(\Proj(y))$ equals $0$
for $y\le0$, $cy$ on $[0,1]$ and $c$ for $y\ge1$. It is convex at $y=0$
but concave at $y=1$. Hence the descent analysis of
Section~\ref{subsec:static} does not apply to the Gaussian variant.

The unit-sphere construction avoids both difficulties.
\end{remark}


\begin{remark}\label{rem:onepoint}
Consider the one-point oracle
\(
  \hat g^{\,1}_\mu(x)\defe\frac{n(b-a)}{\mu}\,f_k(S_\tau(x+\mu u))\,u,
\)
which discards the base evaluation in \eqref{eq:oracle}.  {It} requires a
single play per iteration, and is therefore implementable in the strict
one-query bandit model.  {All plays are then} counted in the regret, and
$\{f_{k^+}\}_{k=0}^N$ is no longer needed.
 {Unbiasedness is unaffected, since the base evaluation contributes nothing
to the mean by $\E_u[u]=0$, see \eqref{eq:pf_cv}. Hence}
$\E_{u,\tau}[\hat g^{\,1}_\mu(x)]=\nabla f^L_{\mu,k}(x)$ by the argument of
 {the proof of} Lemma~\ref{lem:unbiased} verbatim. For the second moment,
$\norm{u}=1$ and $|f_k(S)|\le M$ give
\(
  \E_{u,\tau}\big[\norm{\hat g^{\,1}_\mu(x)}^2\big]
  \;\le\;\frac{M^2n^2(1+2\mu)^2}{\mu^2}.
\)
Item~(i) of Theorems~\ref{thm:static} and~\ref{thm:dynamic} therefore holds
for $\hat g^{\,1}_\mu$ with a similar proof  {and} slightly different
constants.  {However, the one-point oracle} discards the coupling induced
by the shared threshold, which is exactly what the base evaluation exploits.
\end{remark}

\begin{remark}[Counting both plays]\label{rem:charge_both}
Theorems~\ref{thm:static} and~\ref{thm:dynamic} count 
only the exploration
play. Suppose the base play is counted as well, with the static and dynamic
comparators $\min_{S}\sum_{k=0}^N\big(f_k(S)+f_{k^+}(S)\big)$ and
$\sum_{k=0}^N\big(\min_S f_k(S)+\min_S f_{k^+}(S)\big)$, and let the
assumption of item~(ii) hold. By Remark~\ref{rem:extended},
$(1+2\mu)\E_{\tau_k}[f_{k^+}(S_{\tau_k}(x_k))]=f^L_{k^+}(x_k)+\mu f_{k^+}([n])$.
Moreover, $|f^L_{k^+}(x)-f^L_k(x)|\le V$ for all $x\in[0,1]^n$ by
Remark~\ref{rem:lovae}, and $\min_S f_{k^+}(S)\ge \min_S f_k(S)-V$. Since the
descent inequalities in the proofs hold for any fixed comparator in $[0,1]^n$,
each base play costs at most the corresponding gap of $f^L_k {(\cdot)}$ plus
$2V+3\mu M$. With the tuning of  {parameters in} item~(ii) of Theorem~\ref{thm:static}, the
regret with both plays counted satisfies
\begin{equation*}
  \E\big[\Regret_N\big]\le\big(3n\bar L_0+6Mn+8\sqrt3\,Mn^{5/4}\big)(N+1)^{2/3}
  +2V(N+1).
\end{equation*}
 {Item}
~(ii) of Theorem~\ref{thm:dynamic}  {can be} 
modified in the same way, with
$n(3\bar L_0+6M)+8\sqrt3\,Mn^{3/4}\sqrt{D_N}$ in place of the constant.
Hence the $O(N^{2/3})$ rates are preserved if, in addition, $V=O(N^{-1/3})$.
\end{remark}

\end{document}